\documentclass[10pt,twocolumn,letterpaper]{article}

\usepackage{cvpr}
\usepackage{times}
\usepackage{epsfig}
\usepackage{graphicx}
\usepackage{amsmath}
\usepackage{amssymb}

\usepackage{multirow}

\usepackage[caption=false]{subfig}
\usepackage{caption} 
\usepackage[breaklinks=true,bookmarks=false]{hyperref}

\cvprfinalcopy 

\def\cvprPaperID{****} 

\newcommand{\revise}[1]{\textcolor{black}{#1}}

\begin{document}

\title{Rotation-Sensitive Regression for Oriented Scene Text Detection}

\author{Minghui Liao\textsuperscript{1}, Zhen Zhu\textsuperscript{1}, Baoguang Shi\textsuperscript{1}, Gui-song Xia\textsuperscript{2}, Xiang Bai\textsuperscript{1}\thanks{Corresponding author.}\\
\textsuperscript{1}Huazhong University of Science and Technology\\ 
\textsuperscript{2}Wuhan University\\ 
{\tt\small \{mhliao, zzhu\}@hust.edu.cn, shibaoguang@gmail.com, guisong.xia@whu.edu.cn, xbai@hust.edu.cn}}

\maketitle

\begin{abstract}
Text in natural images is of arbitrary orientations, requiring detection in terms of oriented bounding boxes.
Normally, a multi-oriented text detector often involves two key tasks: 1) text presence detection, which is a classification problem disregarding text orientation; 2) oriented bounding box regression, which concerns about text orientation. Previous methods rely on shared features for both tasks, resulting in degraded performance due to the incompatibility of the two tasks.
To address this issue, we propose to perform classification and regression on features of different characteristics, extracted by two network branches of different designs.
Concretely, the regression branch extracts rotation-sensitive features by actively rotating the convolutional filters, while the classification branch extracts rotation-invariant features by pooling the rotation-sensitive features. The proposed method named Rotation-sensitive Regression Detector (RRD) achieves state-of-the-art performance on three oriented scene text benchmark datasets, including ICDAR 2015, MSRA-TD500, RCTW-17 and COCO-Text.
Furthermore, RRD achieves a significant improvement on a ship collection dataset, demonstrating its generality on oriented object detection.

\end{abstract}
\section{Introduction} 
Reading text in the wild is an active research field in computer vision, driven by many real-world applications such as license plate recognition~\cite{licenseplate}, guide board recognition~\cite{guideboard}, and photo OCR~\cite{photoocr}. A scene text reading system generally begins with localizing text regions on which the recognition is then performed. Consequently, one of the main bottleneck of such a system lies in the quality of text detection. 

\begin{figure}[htbp]
\begin{center}
\subfloat[Input image\label{subfig-1:region_and_feature}]{%
       \includegraphics[width=0.13\textwidth]{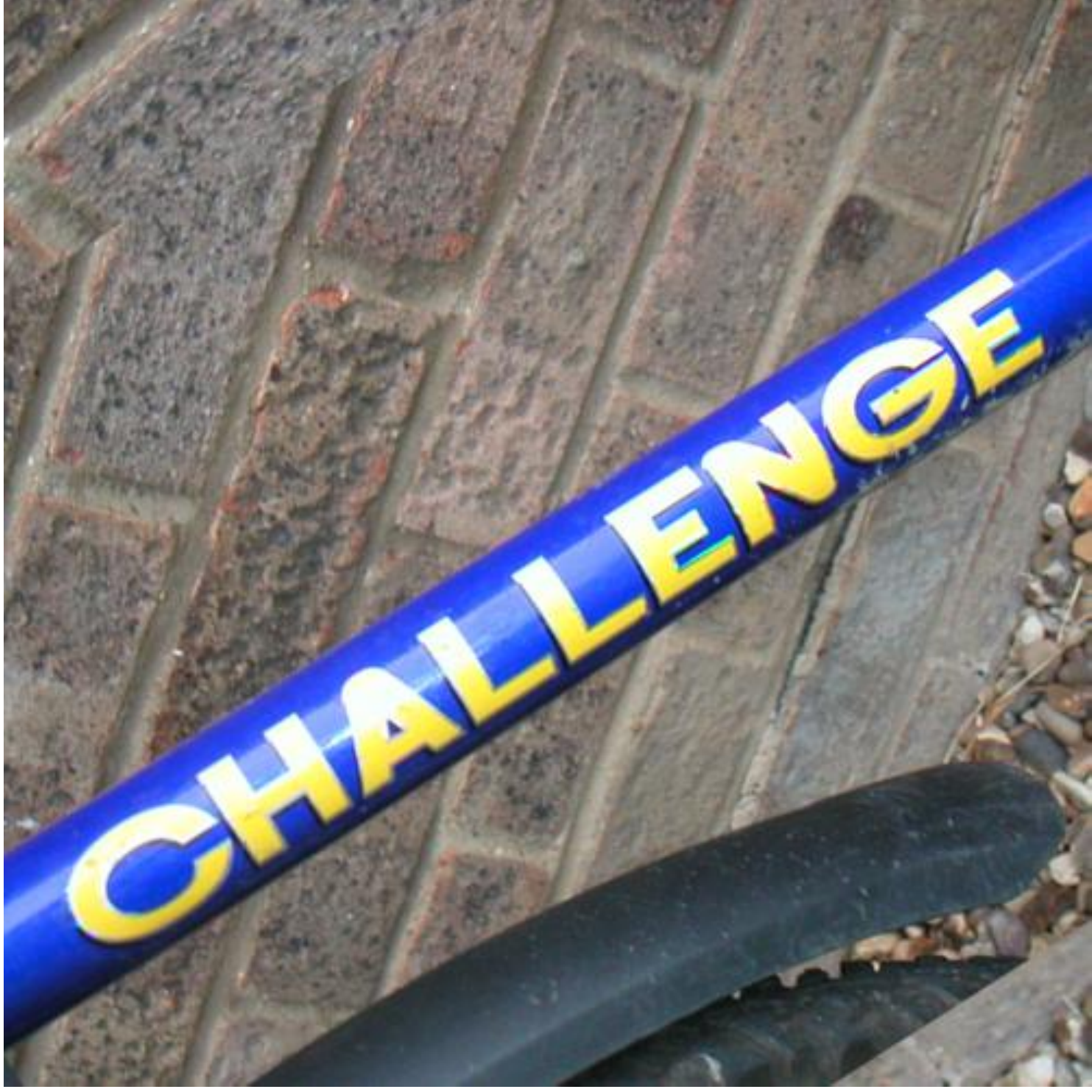}
     }
\subfloat[Shared feature\label{subfig-2:region_and_feature}]{%
       \includegraphics[width=0.13\textwidth]{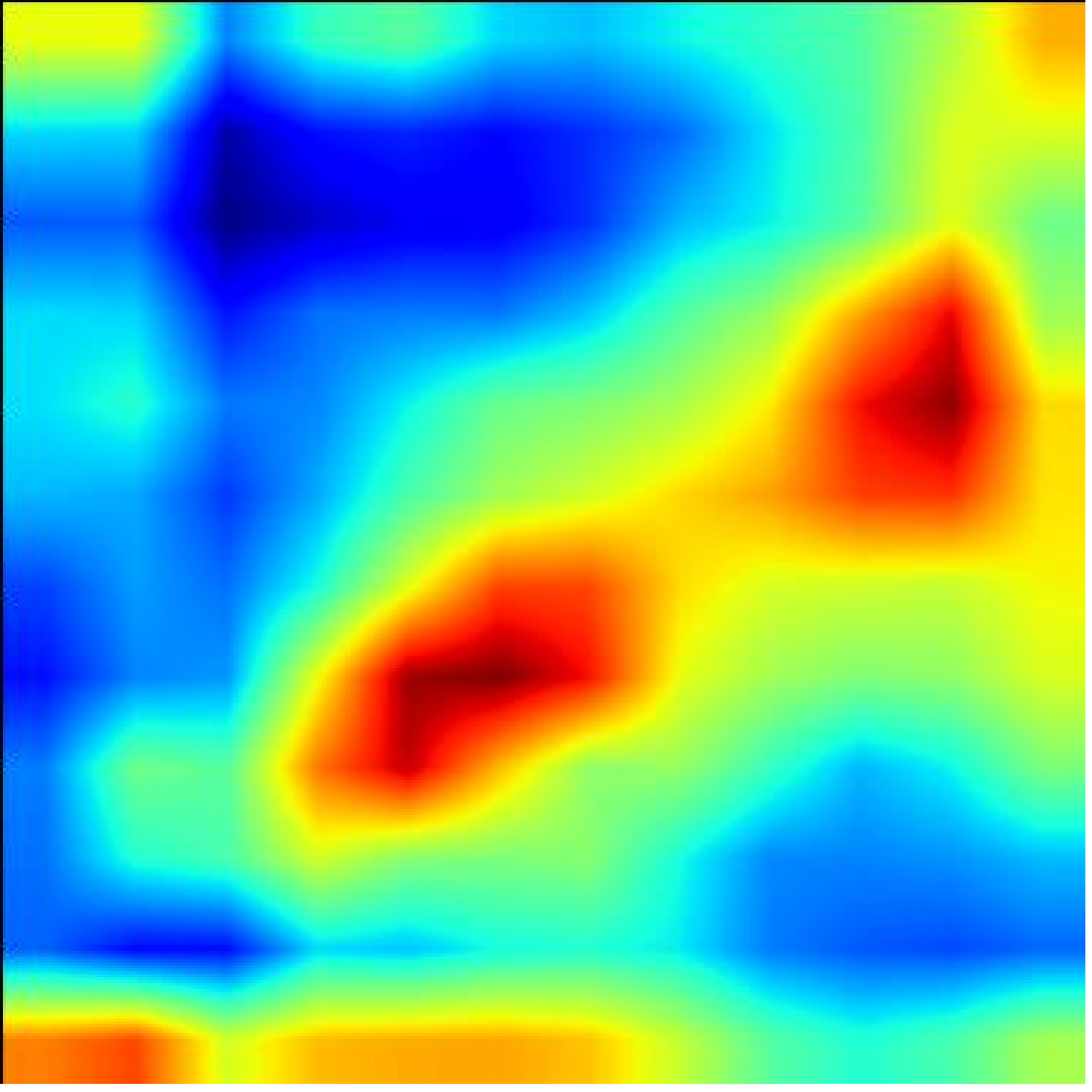}
     }
\subfloat[Result\label{subfig-3:region_and_feature}]{%
       \includegraphics[width=0.13\textwidth]{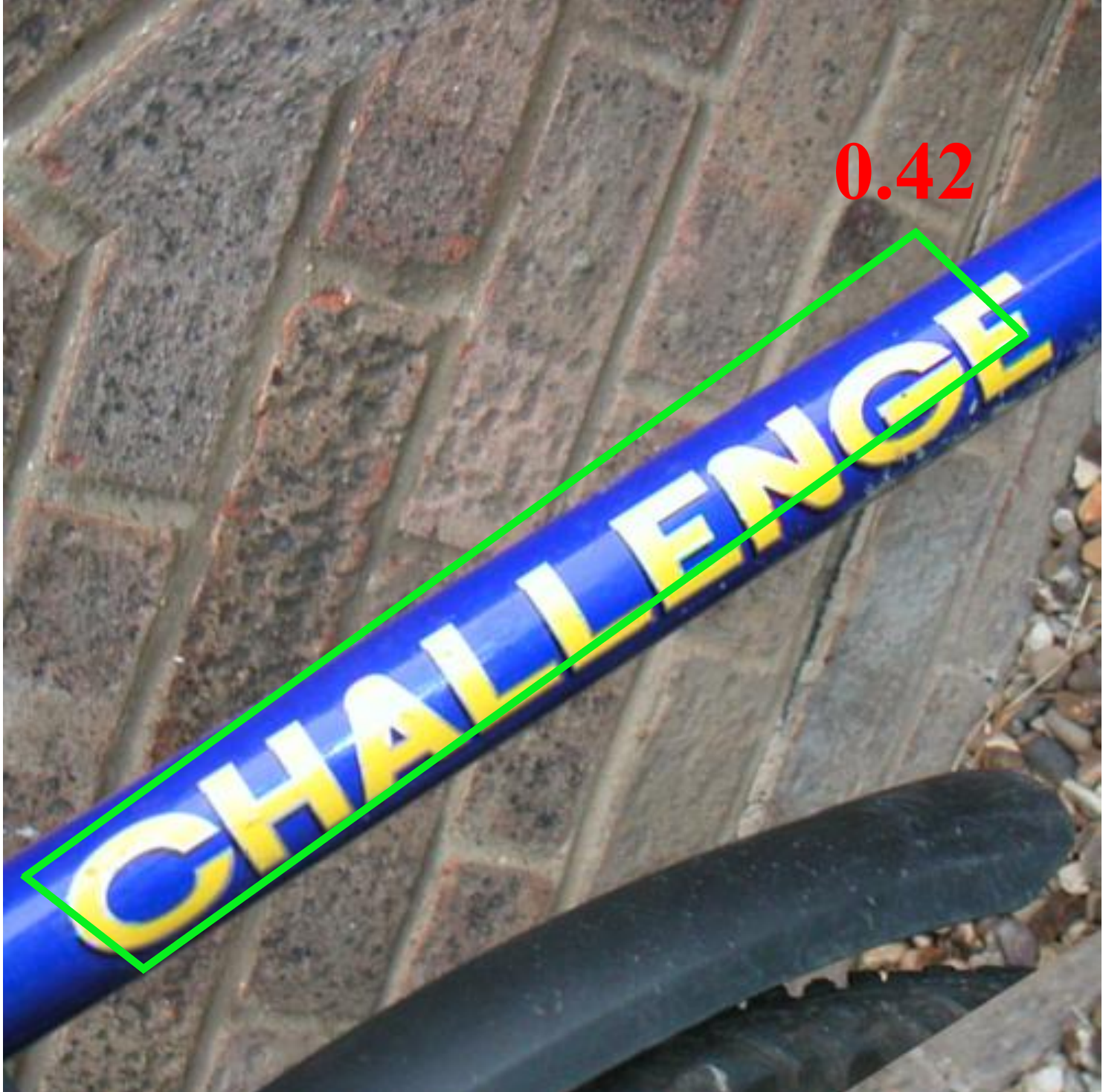}
     }
\vspace{-3mm}
\hfill
\subfloat[Reg. feature\label{subfig-4:region_and_feature}]{%
       \includegraphics[width=0.13\textwidth]{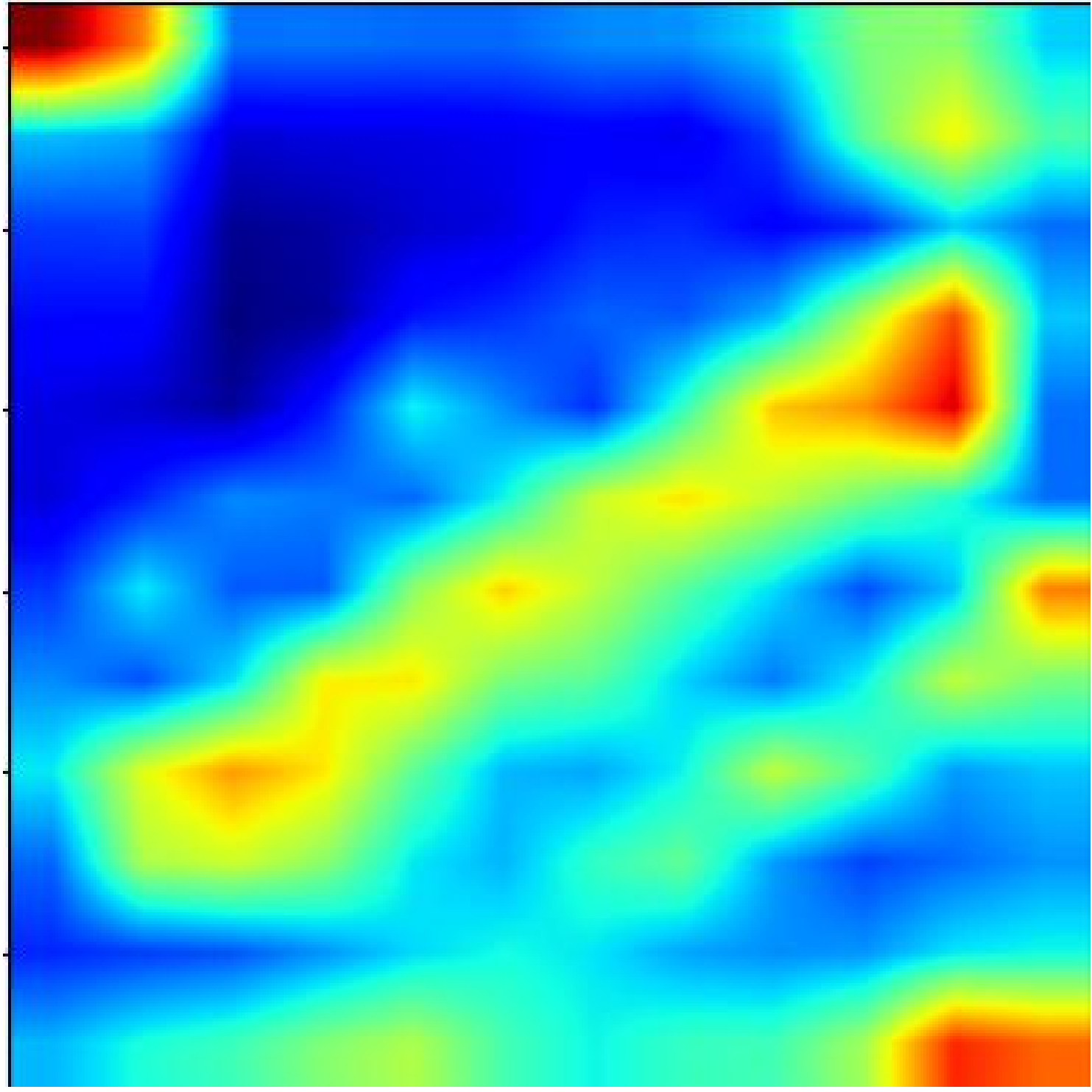}
     }
\subfloat[Cls. feature\label{subfig-5:region_and_feature}]{%
       \includegraphics[width=0.13\textwidth]{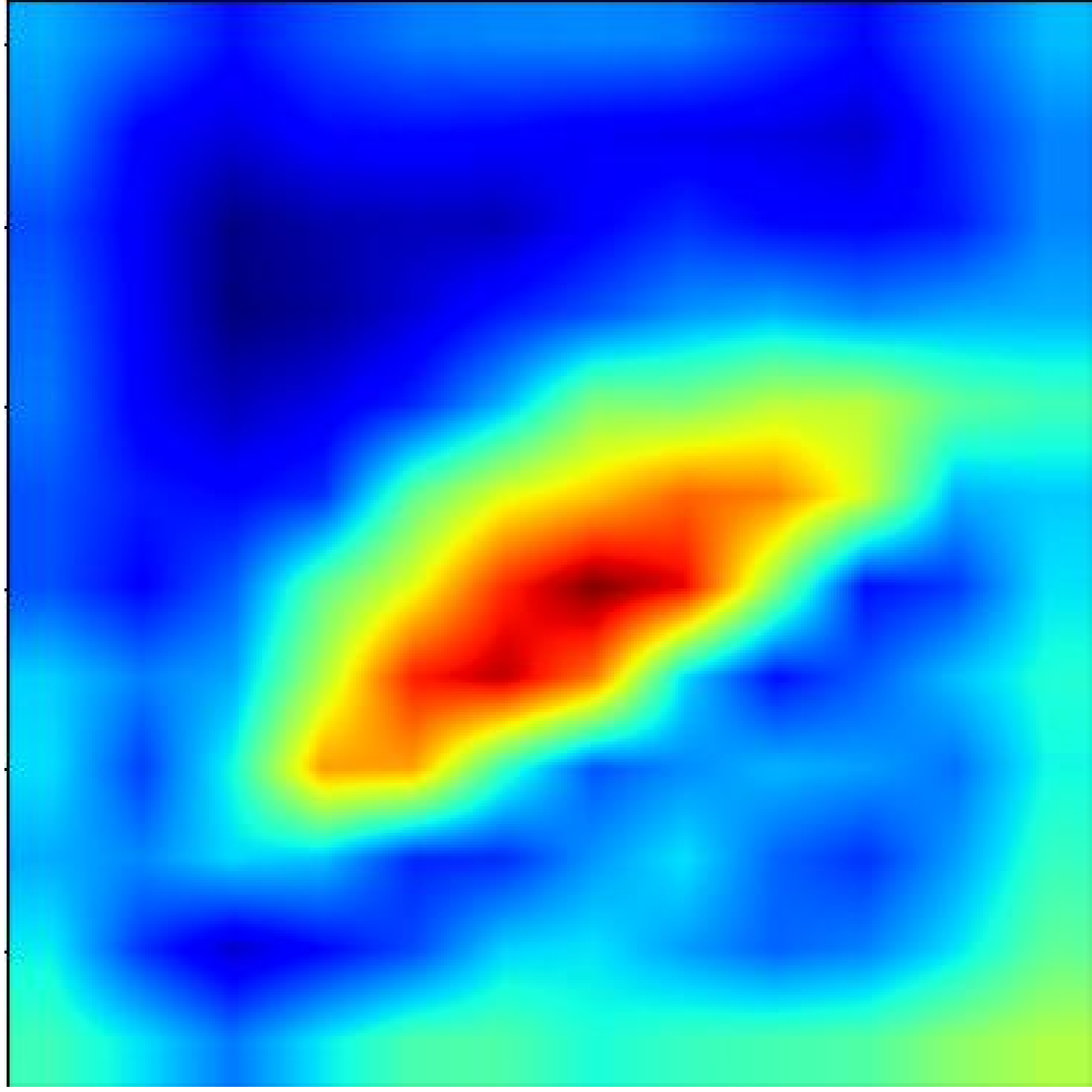}
     }
\subfloat[Result\label{subfig-6:region_and_feature}]{%
       \includegraphics[width=0.13\textwidth]{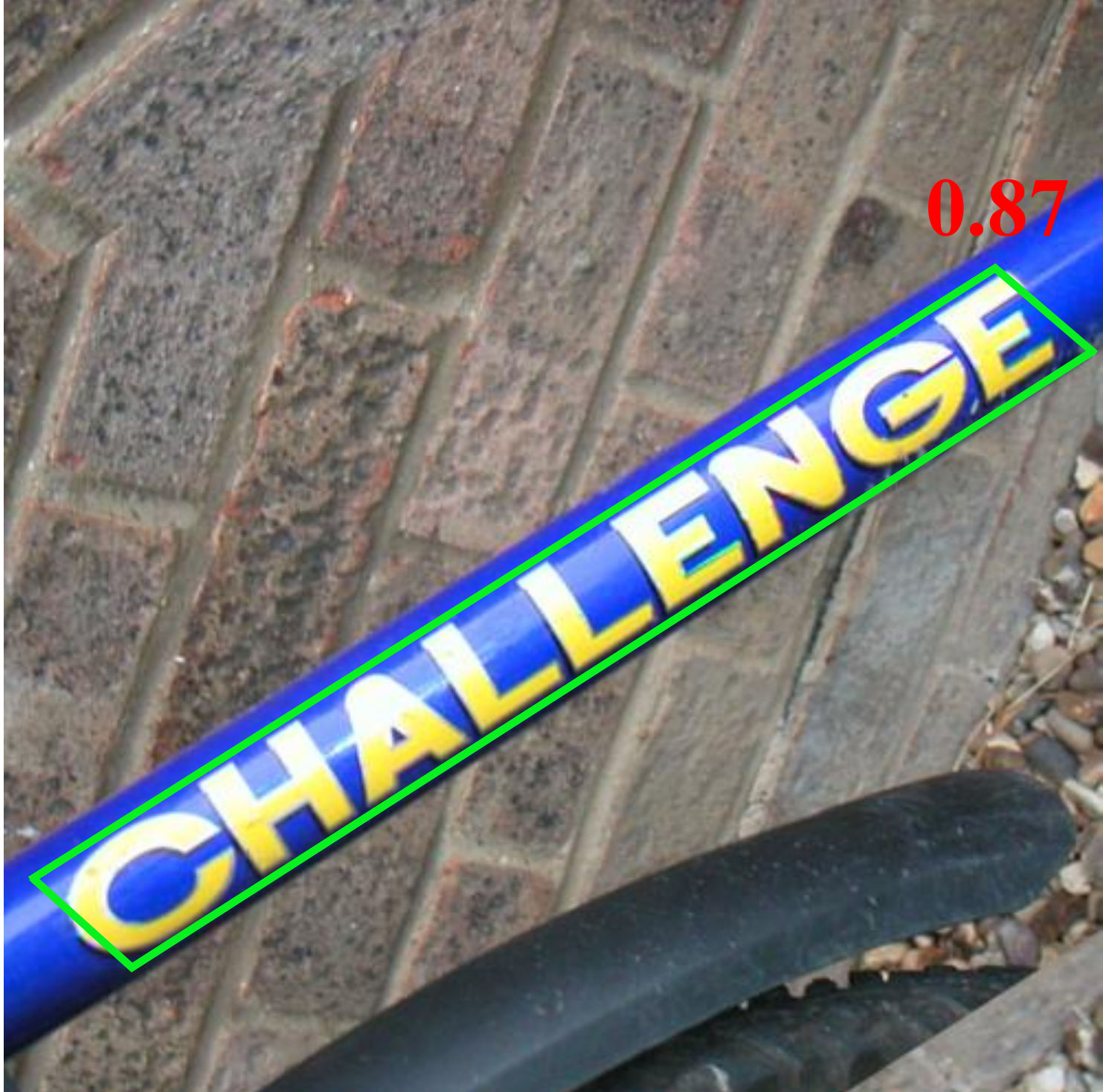}
     }
\end{center}
\vspace{-2mm}
\caption{Visualization of feature maps and results of baseline and RRD. Red numbers are the classification scores. (b): the shared feature map for both regression and classification; (c): the result of shared feature; (d) and (e): the regression feature map and classification feature map of RRD; (f): the result of RRD. }
\label{fig:region_and_feature}
\vspace{-6mm}
\end{figure}

Despite the great success of recent general object detection algorithms~\cite{rcnn,fast_rcnn,ren2015faster,yolo,liu2015ssd,yolo9000}, scene text detection remains challenging mainly due to arbitrary orientations, small sizes, and significantly varied aspect ratios of text in natural images. In fact, general object detection methods usually focus on detecting objects in terms of horizontal bounding boxes, which are accurate enough for most objects such as person, vehicles, etc. Yet, horizontal bounding box is not appropriate for representing long and thin objects in arbitrary orientations (see Fig.~\ref{subfig-1:region_and_feature}). Text in natural images is a typical example of multi-oriented long and thin object, which is better covered by oriented bounding boxes. Directly applying general object detection methods to scene text detection would generally lead to poor performance. 

Some recent scene text detectors~\cite{LiaoSBWL17, deepmatch} successfully adopt general object detection methods to scene text detection with some dedicated designs, yielding a great improvement in scene text detection. Generally, recent object detection consists of predicting object category ({\em i.e.,} classification) and regressing bounding box for accurate localization. Both tasks rely on shared features which are rotation-invariant attributing to the use of pooling layers in general convolutional neural network (CNN) architecture~\cite{BoureauPL10,LencV15,Goodfellow-et-al-2016-Book}. This pipeline is also adopted in recent scene text detectors inspired by general object detection methods. Although it is well known that rotation-invariant features can boost the performance of classification, the rotation invariance is not beneficial for regressing arbitrary oriented bounding boxes. This conflicting issue between classification and regression may not be very important on oriented objects with limited aspect ratio.
\revise{However, unlike Latin text, there is not a ``blank" between neighbor words in non-Latin text such as Chinese, Arabic, Japanese, etc, which possess long text lines frequently and are often detected at the line level instead of word spotting. In this sense, detecting long oriented text lines is obviously a non-trivial task that satisfies with the practical requirements of a more general text reading system for multi-lingual text.}
Thus, for scene text, especially non-Latin text lines which are usually long and thin, and of arbitrary orientations, using rotation-invariant features would hinder the regression of such oriented bounding boxes. 
Another important issue for detecting arbitrary oriented long text having extreme aspect ratios is the requirement of more flexible receptive field. 

\begin{figure*}[ht]
\centering
\subfloat[Backbone\label{subfig-1:architecture}]{%
       \includegraphics[width=0.38\textwidth]{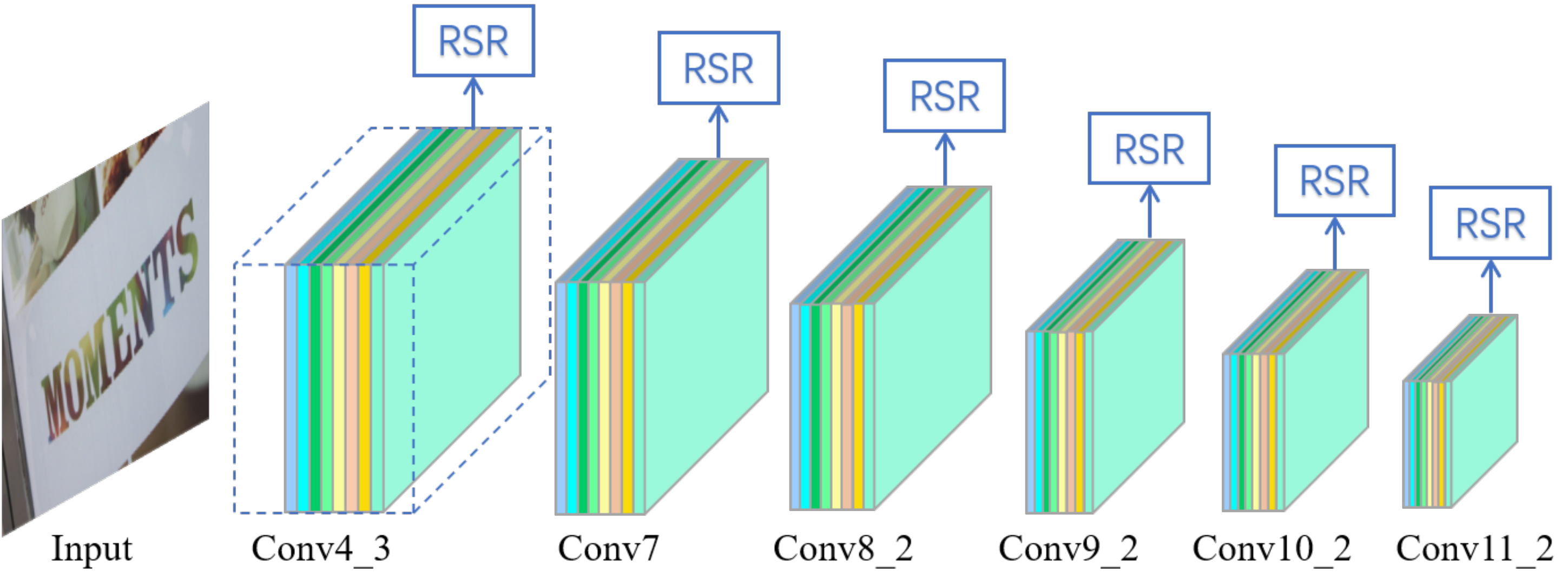}
     }
\hspace*{\fill}
\subfloat[Rotation-Sensitive Regression (RSR)\label{subfig-2:architecture}]{%
       \includegraphics[width=0.55\textwidth]{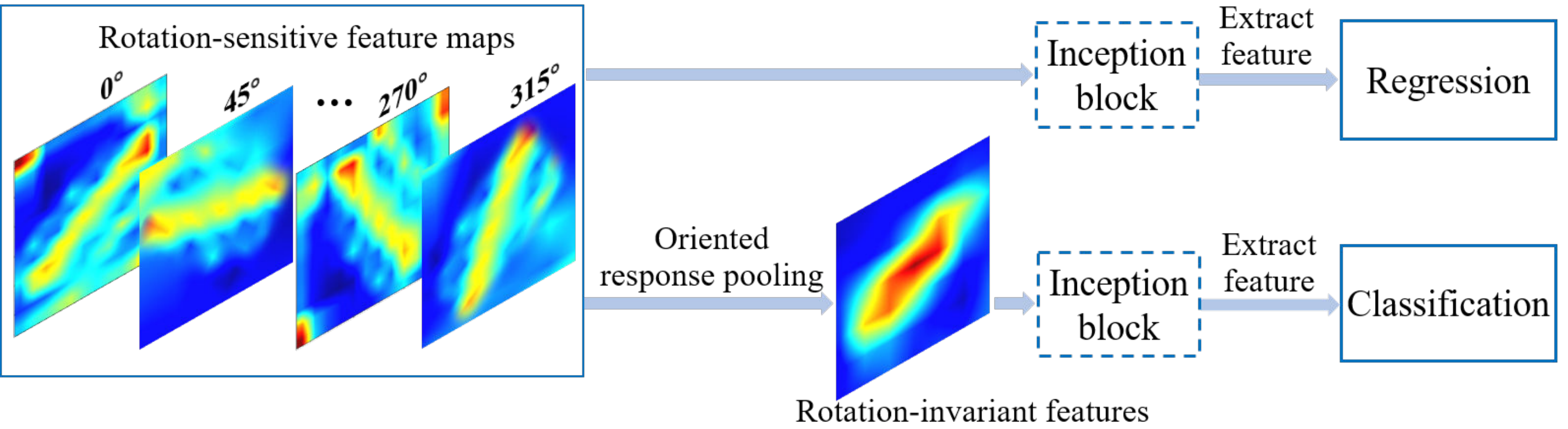}
     }
\caption{Architecture of RRD. (a) The rotation-sensitive backbone follows the main architecture of SSD while changing its convolution into oriented response convolution. (b) The outputs of rotation-sensitive backbone are rotation-sensitive feature maps, followed by two branches: one for regression and another for classification based on oriented response pooling. Note that the inception block is optional.}
\label{fig:architecture}
\vspace{-6mm}
\end{figure*}

To alleviate the above-mentioned issues of arbitrary oriented scene text detection, we propose to separate the regression task from the classification task. More specifically, as depicted in Fig.~\ref{fig:architecture}, the proposed method named Rotation-sensitive Regression Detector (RRD) performs classification with rotation-invariant features, and oriented bounding box regression with rotation-sensitive features. For that, we adopt oriented response convolution~\cite{orn} instead of normal convolution in the network architecture. The rotation-invariant features for classification are then obtained by an oriented response pooling layer~\cite{orn}. To the best of our knowledge, this is the first time to apply oriented response network to object detection task. In addition, we also propose an inception block of three-scale convolutional kernels to give flexible receptive field better covering long text.
As shown in Fig.~\ref{fig:region_and_feature}, the regression feature map (Fig.~\ref{subfig-4:region_and_feature}) of RRD contains richer and more precise orientation information and its classification feature map (Fig.~\ref{subfig-5:region_and_feature}) is more intensive, compared with the conventional shared feature map (Fig.~\ref{subfig-2:region_and_feature}) for both classification and regression. The final results (Fig.~\ref{subfig-3:region_and_feature} and Fig.~\ref{subfig-6:region_and_feature}) also demonstrate the quality of the feature maps.

The main contributions of this paper are three folds: 1) We propose a novel idea of using rotation-sensitive features for regressing oriented bounding boxes while using rotation-invariant features for classification. This separation gives rise to a more accurate regression in detecting arbitrary oriented long and thin objects; 2) A general framework for arbitrary oriented object ({\em e.g.}, scene text) detection is proposed. It can be easily embedded into any existing detection architectures, improving the performance without obvious loss of speed. 
3) The proposed RRD is also an effective and efficient oriented scene text detector, with the generality of detecting both Latin and non-Latin text.

\section{Related Work} \label{sec:related works}
\subsection{Object Detection} \label{subsec:object detection}

Recent object detectors~\cite{rcnn,fast_rcnn,ren2015faster,yolo,liu2015ssd, rfcn,yolo9000,fpn} leverage the powerful learning ability of CNN to detect objects, and achieve impressive performances. Generally, most CNN-based object detectors share a common pipeline which consists of object classification predicting the object category and bounding box regression for accurate localization. Both classification and regression rely on shared translation and rotation-invariant features attributing to pooling layers involved in classical CNN architecture. Although these invariances are beneficial and important for classification task, accurate regression requires translation and rotation-sensitive features. In this sense, classification and regression are somewhat incompatible to each other about the demands of translation invariance~\cite{rfcn} and rotation invariance. R-FCN~\cite{rfcn} introduces sensitive ROI-pooling to alleviate the translation invariance problem. In~\cite{yolo9000, fpn}, the authors propose to fuse the high-resolution and low-resolution feature maps to balance the conflict. To the best of our knowledge, the rotation invariance incompatibility has never been explicitly considered for object detection. This is particularly important for regressing oriented bounding boxes, which are more appropriate for detecting arbitrary oriented long and thin objects.  This paper focuses on a typical example of such object detection, i.e., scene text detection, by explicitly introducing rotation-sensitive features to the CNN pipeline. 

\subsection{Scene Text Detection} \label{subsec:scene text detection}
Different from scene text proposal methods~\cite{DK_textproposal,DK_fast} which mainly concern the recall, scene text detection methods output much less output bounding boxes, considering the trade-off between recall and precision.
Recently, numerous inspiring ideas and promising methods~\cite{MSRA,zhang2015symmetry,textflow,jaderberg2016reading,Zhang_2016_CVPR,SynthText,eccv/TianHHH016,LiaoSBWL17,east, seglink,deepmatch,deepdirect,He_2017_ICCV,Hu_2017_ICCV,Wu_2017_ICCV,Busta_2017_ICCV,Li_2017_ICCV,Tian_2017_ICCV} have been proposed. A great improvement has been achieved compared to traditional methods~\cite{huang2013Text,matas2004robust,huang2014robust}. Based on the representation of detection output, scene text detectors could be roughly divided into two categories: 1) Horizontal text detectors~\cite{zhang2015symmetry,jaderberg2016reading,SynthText,SynthText,eccv/TianHHH016,LiaoSBWL17} which detect words or text lines in terms of horizontal bounding boxes in the same way as general object detectors.
This implies that horizontal scene text detection may benefit from the developments of general object detection such as~\cite{yolo,liu2015ssd}. The works in~\cite{SynthText,LiaoSBWL17} are such examples; 2) Multi-oriented text detectors~\cite{MSRA,Zhang_2016_CVPR,east,seglink,deepmatch,deepdirect,He_2017_ICCV,Hu_2017_ICCV,Wu_2017_ICCV} which focus on detecting text of arbitrary orientations. The detection outputs are usually represented by either oriented rectangles or more generally quadrilaterals enclosing arbitrary oriented words or text lines. Compared to horizontal bounding box representation, additional
variables such as the angle or vertex coordinates are required for representing multi-oriented text bounding boxes.

All the modern scene text detectors inspired by recent CNN-based general object detectors use shared features for both classification and regression. Compared to these modern multi-oriented scene text detectors, this paper proposes to explicitly use rotation-sensitive CNN features for oriented bounding box regression while adopting rotation-invariant features for classification. This results in a more accurate oriented bounding box regression for arbitrary oriented text, especially for the long words or text lines. 

\subsection{Rotation-sensitive CNN Features} \label{subsec:rotation-sensitive CNN features}
Rotation-invariant features are important for a robust classification. Modern CNN architectures usually involve pooling layers achieving rotation invariance to a certain extent. Some recent works~\cite{rotinvar1,rotinvar2,orn} focus on enhancing the rotation invariance of the CNN features to further improve the classification performance. For example, ORN~\cite{orn} proposes to actively rotate during convolution, producing rotation-sensitive feature maps. This is followed by an oriented pooling operation, giving rise to enhanced rotation invariance, and thus resulting in better classification performance. The proposed RRD is inspired by ORN~\cite{orn}. The enhanced rotation-invariant features are used for text presence prediction. We propose to adopt rotation-sensitive features to regress oriented bounding boxes, yielding accurate detection of arbitrary oriented long objects ({\em e.g.}, scene text). To the best of our knowledge, this is the first time that explicit rotation-sensitive features are used for arbitrary oriented object detection. 

\section{Rotation-Sensitive Regression Detector} \label{sec:method description}
\subsection{Overview}

RRD is an end-to-end trainable, fully convolutional neural network whose architecture is inspired by SSD~\cite{liu2015ssd}.
Its architecture is illustrated in Fig.~\ref{fig:architecture}, which uses VGG16~\cite{simonyan2014very} as its backbone network, with extra layers added in the same manner as SSD.
As shown in Fig.~\ref{fig:architecture}, six layers of the backbone network are taken for dense prediction.

The dense prediction is similar to that of SSD~\cite{liu2015ssd}. 
For every default box~\cite{liu2015ssd}, RRD classifies its label (text or non-text) and regresses relative offsets.
After that, RRD applies the offsets to the default boxes classified as positive, producing a number of quadrilaterals, each with a score.
The quadrilaterals are filtered by non-maximum suppression, which outputs final detections.

The key novelty of our method is the dense prediction part, where RRD extracts two types of feature maps of different characteristics for classification and regression respectively.
The feature map for classification is insensitive to text orientation, while the feature map for regression is sensitive. As mentioned before, these characteristics well fit the nature of the two tasks.

\subsection{Rotation-Sensitive Regression}
Text coordinates are sensitive to text orientation.
Therefore, the regression of coordinate offsets should be performed on rotation-sensitive features.
Oriented response convolution encodes the rotation information by actively rotating its convolutional filters, producing rotation-sensitive features for regression.

Different from standard CNN features, RRD extracts rotation-sensitive features with active rotating filters (ARF)~\cite{orn}.
An ARF convolves a feature map with a \emph{canonical filter} and its rotated clones. The filters are rotated following the method in~\cite{orn}.
Denote the canonical filter of ARF as $F_0 \in \Re^{k\times k\times N}$, where $k$ is the kernel size, $N$ is the number of rotations.
ARF makes $N-1$ clones of the canonical filter by rotating it to different angles, respectively $F_j, {j=1:N}$. Let $M_i(j)$ and $M_o(j)$ denote the input feature map and the output feature map of $j$-th orientation respectively.
ARF convolves a feature map by computing:
\begin{equation}
M_o(j)=\sum_{n = 0}^{N-1}F_j(n)*M_i(n), j=0,...,N-1,
\end{equation}
where $F_j(n)$ indicates the n-th orientation channel of $F_j$.
After convolution, it produces a response map of $N$ channels, each corresponding to the response of the canonical filter or its rotated clone.
$N$ is set to 8 in practice.

ARF produces extra channels to incorporate richer rotation information.
With the help of ARF, ORN produces feature maps with orientation channels, capturing rotation-sensitive features and improving its generality for rotated samples which has never seen before.
Besides, since the parameters between the $N$ filters are shared, learning ARF requires much less training examples.

In addition, in order to make the receptive field suitable for long text lines, we adopt inception blocks for both branches.
The inception block concatenates the output feature maps produced by three filters of different sizes.
The filter sizes are respectively $m\times m$, $m\times n$ and $n\times m$, where $m$ is set to 3 and $n$ is set to \((9,7,5)\) in the first, the second, and the last stages respectively.
Inception blocks result in receptive fields of different aspect ratios.
They are particularly helpful for detecting long text.
Therefore, they are used in line-based text detection, but discarded in word-based text detection, where long text is rare.

\subsection{Rotation-Invariant Classification}

In contrast to regression, the classification of text presence should be rotation-invariant, \ie, text regions of arbitrary orientations should be classified as positive.
Therefore, a rotation-invariant feature map should be extracted for this task.

ORN achieves rotation invariance by pooling responses of all $N$ response maps.
As shown in Fig.~\ref{subfig-2:architecture}, the rotation-sensitive feature maps are pooled along their depth axis.
Assuming that $M_{or}$ is a rotation-sensitive input feature map of \(N\) orientation channels, the rotation-invariant feature map \(M_{pooling}\) is an element-wise max of $M_{or}$ with the index of orientations, which can be calculated as follows:
\begin{equation}
M_{pooling}=\max_{k=0}^{N-1}{M_{or}(k)}, 
\end{equation}

Since the pooling operation is orderless and applied to all $N$ response maps, the resulting feature map is locally invariant to object rotation. Therefore, we use this feature map for classification.
Besides, the setting of inception block is the same as the regression branch.

\subsection{Default Boxes and Prediction}
The default boxes are horizontal rectangles with different sizes and aspect ratios.
Let $\mathbf{B}_0=(x_0, y_0, w_0, h_0)$ denote a horizontal default box, which can also be represented by its four vertexes $\mathbf{Q}_0=(v^0_{1}, v^0_{2}, v^0_{3}, v^0_{4})$, where $v^0_{i} = (x^0_{i}, y^0_{i}), i\in\{1,2,3,4\}$.

The regression branch predicts offsets from a default box to a quadrilateral.
A quadrilateral is described as $\mathbf{Q}=(v_1,v_2,v_3,v_4)$, where $v_i=(x_i,y_i), i\in\{1,2,3,4\}$ are four vertexes of a quadrilateral. For each default box, the prediction layer outputs the classification scores and offsets $(\Delta x_1, \Delta y_1, \Delta x_2, \Delta y_2,\Delta x_3, \Delta y_3, \Delta x_4, c)$ between the default box $\mathbf{Q}_0$ and the bounding box result $\mathbf{Q}$. The final output quadrilateral is encoded with the corresponding default box:
\begin{align}
\begin{split}
  x_{i} &= x^0_{i} + w_0\Delta x_{i}, {i = 1, 2, 3, 4}, \\
  y_{i}&= y^0_{i} + h_0\Delta y_{i}, {i = 1, 2, 3, 4},
\end{split}
\label{eq:encode-polygon}
\end{align}
where $w_0$ and $h_0$ denote the width and height of the default box respectively. 
In addition, a non-maximum suppression with quadrilaterals is applied in the prediction period.

\subsection{Training}
\paragraph{Ground Truth.}
The ground truth of an oriented text region can be described as a quadrilateral $\mathbf{G}_q = (v_1, v_2, v_3, v_4)$, where $v_i=(x_i,y_i), i\in\{1,2,3,4\}$ are the vertices of the quadrilateral. We argue that careful selection of the first point in quadrilateral is helpful for regression. Thus, we follow a scheme in \cite{TextBoxes++}, which determine the first point based on the distances from its corresponding maximum horizontal bounding box.
\vspace{-4mm}
\paragraph{Loss Function.}
In the training phase, default boxes are matched to the ground-truth boxes according to  box overlap following the match scheme in~\cite{liu2015ssd}.  For efficiency, the minimum horizontal rectangle enclosing the quadrilateral is used in the matching period.

We adopt a similar loss function to the one used in~\cite{liu2015ssd}. More specifically, let $x$ be the match indication matrix. For the $i$-th default box and the $j$-th ground truth, $x_{ij}=1$ means a match following the box overlap between them, otherwise $x_{ij}=0$. Let $c$ be the confidence, $l$ be the predicted location, and $g$ be the ground-truth location. The loss function is defined as:
\begin{equation}
L(x,c,l,g)=\frac{1}{N}(L_{\textrm{cls}}(x,c)+\alpha L_{\textrm{reg}}(x,l,g)),
\end{equation}
where $N$ indicates the number of default boxes that match ground-truth boxes, and $\alpha$ is set to 0.2 for quick convergence. We adopt a smooth L1 loss~\cite{fast_rcnn} for $L_{\textrm{reg}}$ and a 2-class softmax loss for $L_{\textrm{cls}}$.
We follow the same online hard negative mining strategy as~\cite{liu2015ssd}.

\section{Experiments} \label{sec:experimental results}
\subsection{Datasets} \label{subsec:dataset}
We conduct extensive experiments to verify the effectiveness of RRD in multiple aspects.
Altogether, seven datasets are used in the experiments.
We first evaluate RRD on RCTW-17~\cite{rctw} and MSRA-TD500~\cite{MSRA} to show its effectiveness at detecting long and oriented text.
Both datasets contain many instances of such, and they are annotated in text lines.
To further assess the importance of using rotation-sensitive features for regressing long and oriented boxes, we construct a subset of RCTW-17 by picking text instances with extreme aspect ratios.
COCO-Text dataset~\cite{coco-text/VeitMNMB16} is evaluated to prove the accurate of regression, which provides a evaluation protocol where the IOU threshold is set to $0.75$.
Then we test RRD on ICDAR 2015 incidental text dataset~\cite{icdar15}, which contains oriented English words.
The ICDAR 2013 focused text dataset~\cite{ICDAR2013} is also evaluated, showing its good performance on horizontal text detection.
In the end, in order to show the generality of RRC, we evaluate RRD on HRSC2016~\cite{liu2017high}, a dataset of high resolution ship collection.
\vspace{-4mm}
\paragraph{Reading Chinese Text in the Wild (RCTW-17)} contains 12,000 images taken from streets, screen shots and indoor scenes etc.
The sizes of images range from small to extremely large;
Since Chinese words are not separated by blank spaces, long text lines are common.
\vspace{-4mm}
\paragraph{RCTW-Long} is a sub-dataset drawn from RCTW-17 featuring long text.
The training set of RCTW-Long consists of 1323 images picked from the original training set, and the test set 537 from the original test set.
Specifically, a bounding box is defined as a long box if its aspect ratio is greater than $t$ or less than $1/t$, otherwise a short box.
$t$ is set to 5 for the training set and 7 for the test set.
We only select the images with more long boxes than short boxes for constructing this dataset.
\vspace{-4mm}
\paragraph{MSRA-TD500} contains 500 natural images taken by pocket cameras from indoor and outdoor scenes.
The dataset is divided into 300 training images and 200 test images.
This dataset contains both English and Chinese text.
Compared to RCTW-17, this dataset contains less text instances per image, but larger variance in text orientations.
\vspace{-4mm}
\paragraph{ICDAR 2015 Incidental Text (IC15)} comes from the Challenge 4 of ICDAR 2015 Robust Reading Competition.
Images of this dataset were captured by Google Glasses in streets, shopping malls, etc., in an incidental manner.
Consequently, many images of this dataset are of low resolution, and text is in various orientations.
IC15 contains 1,000 training images and 500 test images.
Annotations are provided in terms of word bounding boxes.
\vspace{-4mm}
\paragraph{COCO-Text} is a large dataset which contains 63686 images, where 43,686 of the images is used for training, 10,000 for validation, and 10,000 for testing. It is one of the challenges of ICDAR 2017 Robust Reading Competition.
\vspace{-4mm}
\paragraph{ICDAR 2013 Focused Text (IC13)} is composed of 229 training images and 233 testing images. This dataset contains only horizontal and focused text. Images in IC13 are in high resolutions.

\vspace{-4mm}
\paragraph{High Resolution Ship Collection 2016 (HRSC2016)} contains 1061 images divided into 436, 181, 444 images for training, validation, test set, respectively. The images are from two scenarios including ships on sea and ships close inshore derived from Google Earth. Ships are abundantly labeled with rotated polygons along with some extra information such as ship types and ship head locations etc. 

\subsection{Implementation Details}
RRD is optimized by the ADAM~\cite{adam} algorithms on all datasets.
For all scene text datasets, RRD is pre-trained on SynthText~\cite{SynthText} for 30k iterations and fine-tuned on real data.
In the first 5--10k iterations, images are resized to $384 \times  384$ after random cropping. The learning rate is fixed to $10^{-4}$.
Another 5--10k iterations is followed, where images are resized to $768 \times  768$ and the learning rate decayed to $10^{-5}$.
The aspect ratios of default boxes are set to $1, 2, 3, 5, 1/2, 1/3, 1/5$ for word-based dataset (IC15 and IC13), and $1, 2, 3, 5,7,9,15, 1/2, 1/3, 1/5, 1/7,1/9,1/15$ for text line-based dataset (RCTW-17, RCTW-Long, and MSRA-TD500).

For HRSC2016 experiments, we scale input images to \(384\times384\) due to lots of thin and long ship instances.
We train the model for around 15k iterations on three NVIDIA TITAN Xp GPUs from scratch at the learning rate of $10^{-4}$ with the batch size set to 32.
Then, we continue training for about 1k iterations with learning rate decayed to $10^{-5}$ while keeping other settings unchanged. The default boxes settings remain the same as the line-based scene text datasets.

\begin{figure*}[ht]
\centering
\includegraphics[width=0.95\linewidth]{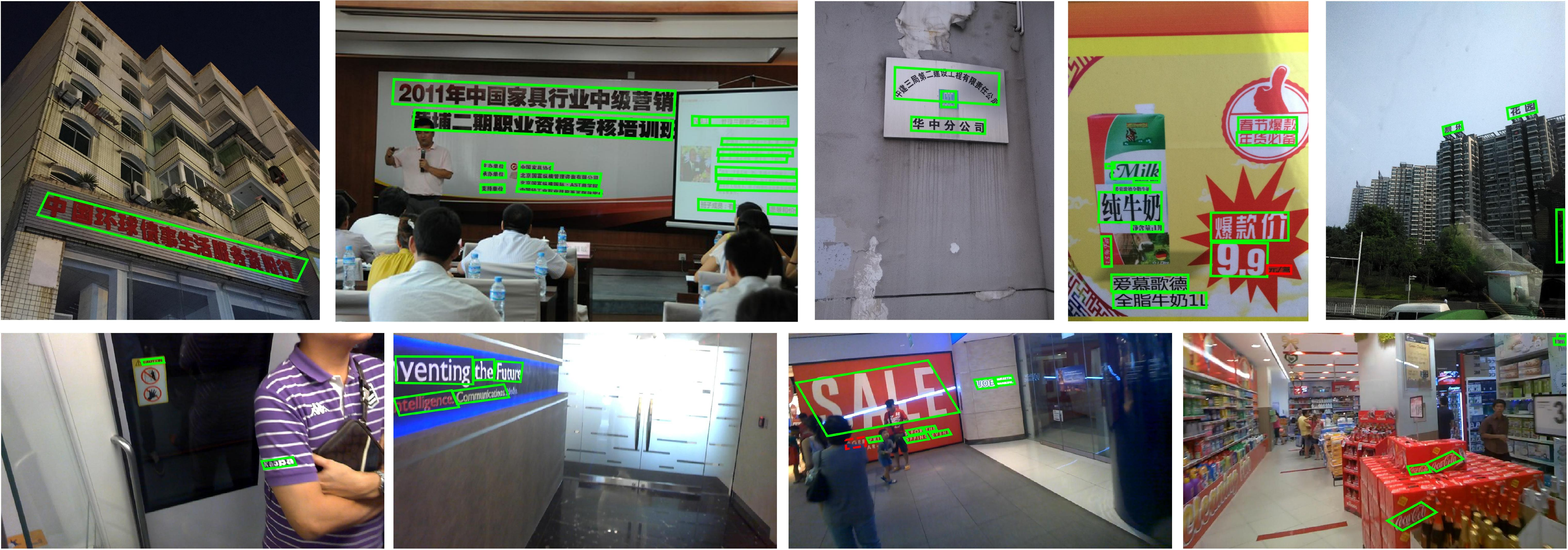}
\caption{Some results of RRD on RCTW-17 (first row) and IC15 (second row).}
\label{fig:text_results_visu}
\vspace{-4mm}
\end{figure*}

\subsection{Ablation Study}
We apply several variants of RRD to verify the effectiveness of rotation-sensitive regression.
The tested variants of RRD are summarized as follows: \\
\textbf{Baseline:} architecture without inception block, using shared conventional feature maps for both regression and classification;
\textbf{Baseline+inc:} baseline architecture using inception blocks;
\textbf{Baseline+inc+rs:} architecture with inception block, using  rotation-sensitive features for both regression and classification;
\textbf{Baseline+inc+rs+rotInvar:} the proposed RRD. Note that for word-based datasets, inception block is not applied and we also name it RRD.

\begin{figure}[htbp]
\begin{center}
\subfloat[Results of Baseline+inc\label{subfig-1:visu_rctw_long}]{%
       \includegraphics[width=0.45\textwidth]{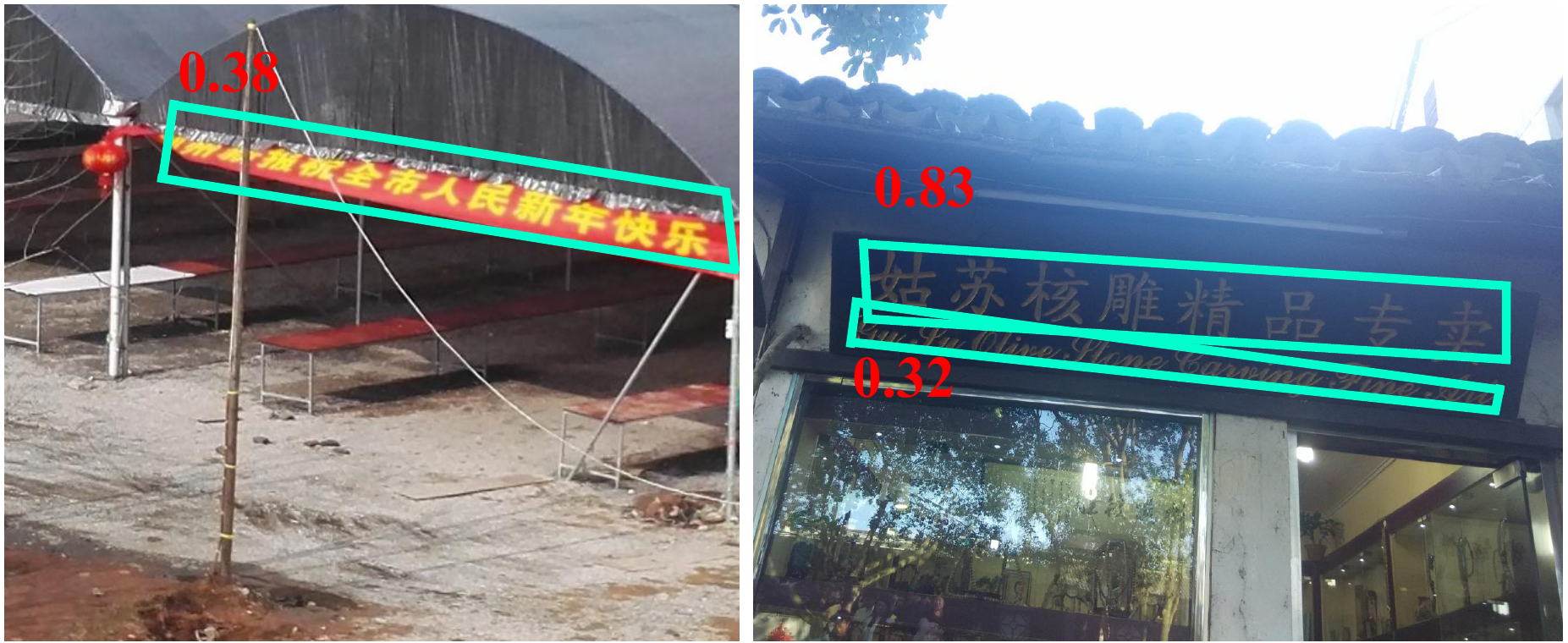}
     }
\vspace{-3mm}
     \hfill
\subfloat[Results of RRD\label{subfig-2:visu_rctw_long}]{%
       \includegraphics[width=0.45\textwidth]{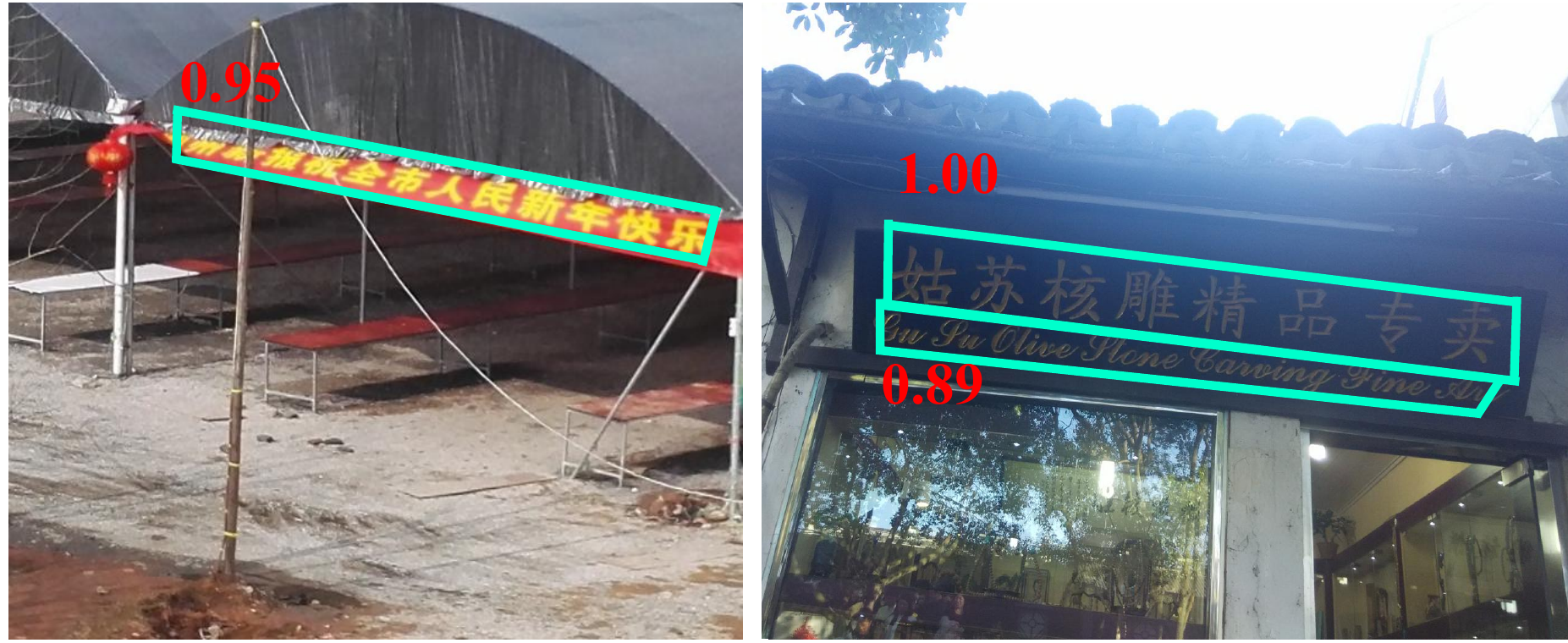}
     }
\end{center}
\vspace{-4mm}
\caption{Some results on images in RCTW-Long dataset. }
\label{fig:visu_rctw_long}
\vspace{-2mm}
\end{figure}

\begin{table}[ht]
\footnotesize
\centering
\begin{tabular}{|c|c|c|c|}
\hline
Method                        & Recall & Precision  & F-measure \\ \hline
Baseline                      & 0.6880 & 0.6925      & 0.6902    \\ \hline
Baseline+inc                  & 0.7102 & 0.8018     & 0.7532    \\ \hline
Baseline+inc+rs           & 0.7423 & \textbf{0.8979}     & 0.8127    \\ \hline
\begin{tabular}[c]{@{}c@{}}Baseline+inc+rs+rotInvar\\ (RRD)\end{tabular} & \textbf{0.8146} & 0.8535     & \textbf{0.8336}    \\ \hline
\end{tabular}
\caption{Evaluation results of several variants of RRD on RCTW-Long dataset. 
\label{table:RCTW-long}
}
\vspace{-4mm}
\end{table}

\begin{figure}
\centering
\includegraphics[width=0.95\linewidth]{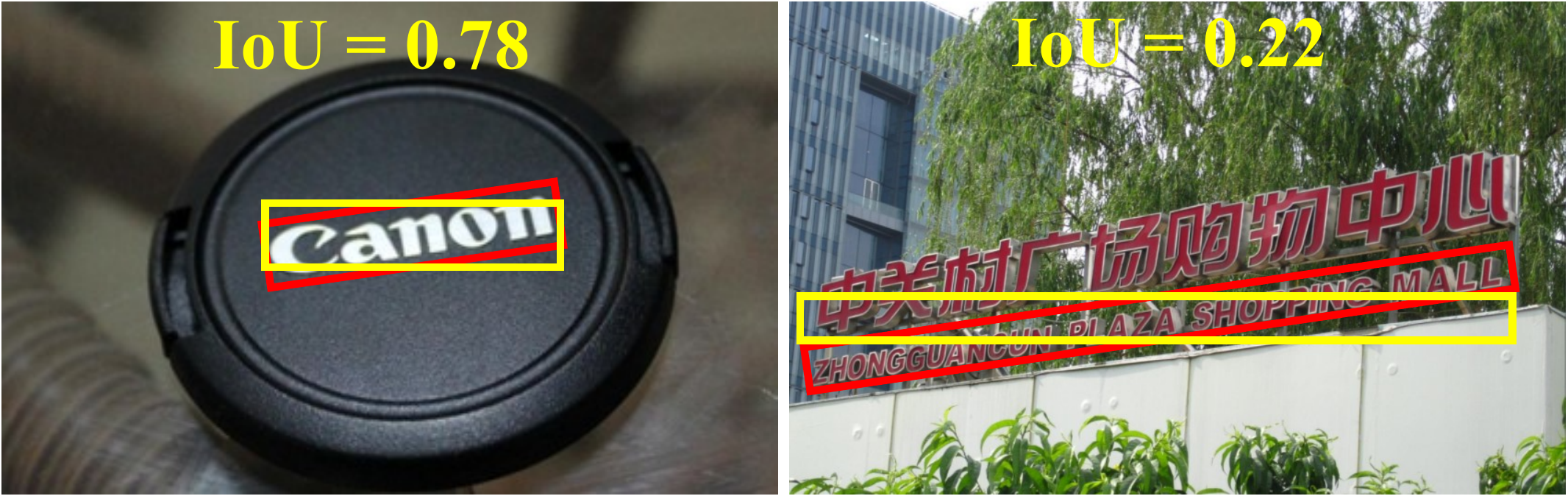}
\caption{Red boxes: ground truths; Yellow boxes: detections. The corresponding angle is the same while the IOU diverges hugely.}
\label{fig:iou_compare}
\vspace{-6mm}
\end{figure}

\revise{Oriented long text needs more accurate bounding boxes.  Given the examples in Fig.~\ref{fig:iou_compare} , the short text can be easily covered by a bounding box even though its orientation is not accurate, while the long text is much more sensitive to the orientation.} Thus, we conduct ablation study on RCTW-Long dataset, a typical dataset which mainly consists of long and oriented text, to verify the superiority of rotation-sensitive regression. The quantitative comparison of several variants of RRD described above is shown in Tab.~\ref{table:RCTW-long}.  Following standard evaluation protocol on RCTW dataset, we traverse the threshold of detection scores from $0.1$ to $1$ with a step of $0.02$ to get the best F-measure for all models. The detailed comparison is given in the following.
\vspace{-4mm}
\paragraph{Inception block.} Compared with the baseline model, the inception block achieves an improvement of about 6 percents. This implies that the inception block can effectively alleviate the limited receptive field problem for detecting long and thin text.
\vspace{-4mm}
\paragraph{Rotation-sensitive regression and classification.} We also test the architecture using rotation-sensitive features for both regression and classification. Such model ``Baseline+inc+rs'' further improves the ``Baseline+inc'' architecture by about 6 percents. Therefore, it is evident that the rotation-sensitive features are useful for long, thin and oriented text detection.

\begin{table*}[ht]
\footnotesize
\centering
\begin{tabular}{|c|c|c|c|c|c|c|c|c|c|}
\hline
\multicolumn{5}{|c|}{\textbf{RCTW-17}}                                                                                           & \multicolumn{5}{c|}{\textbf{ICDAR2015}}                                                                                           \\ \hline
Methods                                      & Recall           & Precision        & F-measure        & FPS             & Methods                                           & Recall          & Precision        & F-measure        & FPS            \\ \hline
Official baseline~\cite{rctw}         & 0.404            & \textbf{0.76}  & 0.528            & 8.9             & Zhang et al.~\cite{Zhang_2016_CVPR}      & 0.43            & 0.71             & 0.54             & -   \\ \hline
EAST-ResNet*                                 & 0.478            & 0.597            & 0.531            & 7.4             & Tian et al.~\cite{eccv/TianHHH016}         & 0.52            & 0.74             & 0.61             & 7.1            \\ \hline
\textbf{Baseline+inc}                  & \textbf{0.459} & 0.659            & 0.541            & \textbf{10.6} & Shi et al.~\cite{seglink}         & 0.768           & 0.731            & 0.750            & 8.9 \\ \hline
\textbf{RRD}                               & 0.453            & 0.724            & \textbf{0.557} & 10              & Liu et al.~\cite{deepmatch}                & 0.682           & 0.732            & 0.706            & -              \\ \hline
\textbf{Baseline+inc+MS}                 & \textbf{0.595} & 0.744            & 0.661            & -               & Zhou et al.~\cite{east}              & 0.735           & 0.836            & 0.782            & \textbf{13.2}           \\ \hline
\textbf{RRD+MS}                              & 0.591            & \textbf{0.775} & \textbf{0.670} & -               & He et al.~\cite{He_2017_ICCV}            & 0.73            & 0.80             & 0.77             & -              \\ \hline
\multicolumn{5}{|c|}{\textbf{MSRA-TD500}}                                                                               & Hu et al.~\cite{Hu_2017_ICCV}    & 0.77            & 0.793            & 0.782            & -              \\ \hline
Methods                                      & Recall           & Precision        & F-measure        & FPS             & \textbf{Baseline}                             & 0.762           & \textbf{0.871} & 0.813            & 8.5            \\ \hline
Zhang et al.~\cite{Zhang_2016_CVPR} & 0.67             & 0.83             & 0.74             & 0.48            & \textbf{RRD}                                      & \textbf{0.79} & 0.856            & \textbf{0.822} & 6.5            \\ \hline
He et al.~\cite{deepdirect}           & 0.7              & 0.77             & 0.74             & 1.1             & Zhou et al. MS~\cite{east}            & 0.783           & 0.833            & 0.807            & -              \\ \hline
Shi et al.~\cite{seglink}    & 0.7              & 0.86             & 0.77             & 8.9             & Hu et al. MS~\cite{Hu_2017_ICCV} & 0.77            & 0.793            & 0.782            & -              \\ \hline
Zhou et al.~\cite{east}         & 0.67             & \textbf{0.87}  & 0.76             & \textbf{13.2} & He et al. MS~\cite{deepdirect}             & 0.80            & 0.82             & 0.81             & -              \\ \hline
\textbf{Baseline+inc}                    & 0.69             & 0.79             & 0.74             & 10.6            & \textbf{Baseline+MS}                          & 0.785           & 0.878            & 0.828            & -              \\ \hline
\textbf{RRD}                                 & \textbf{0.73}  & \textbf{0.87}  & \textbf{0.79}  & 10              & \textbf{RRD+MS}                                   & \textbf{0.8}  & \textbf{0.88}  & \textbf{0.838} & -              \\ \hline
\end{tabular}
\caption{Text detection results on multi-oriented scene text benchmarks: MSRA-TD500, RCTW-17, and IC15. 
}
\label{tab:text_benchmarks}
\vspace{-4mm}
\end{table*}

\vspace{-4mm}
\paragraph{Rotation-sensitive regression and Rotation-invariant classification.} The last model in Tab.~\ref{table:RCTW-long} is the proposed RRD which uses rotation-sensitive and rotation-invariant features for regression and classification respectively. It alleviates the dilemma between regression and classification, outperforming all other models in Tab~\ref{table:RCTW-long} by a large margin.

Some qualitative comparisons between the proposed RRD and its variant ``Baseline+inc'' are illustrated in Fig~\ref{fig:visu_rctw_long}. As shown, the proposed RRD yields more convincing classification scores and more accurate bounding boxes. Specifically, the bounding boxes having scores lower than 0.5 are discarded in practice. Consequently, some text bounding boxes generated by ``Baseline+inc'' are discarded. Whereas, all text bounding boxes given by RRD are reserved. This qualitative comparison also shows the effectiveness of RRD thanks to the specialized features for regression and classification, alleviating the incompatibility of the two tasks. 

\subsection{Results on Scene Text Benchmarks}
\paragraph{RCTW-17.}
RCTW-17 is a large line-based dataset which mainly consists of Chinese text. Thus, the inception block is important. The quantitative results using the official evaluation scheme is depicted in Tab.~\ref{tab:text_benchmarks}. RRD outperforms the official baseline based on SegLink~\cite{seglink} by about 3.3 percents in terms of F-measure. Even though EAST-ResNet~\footnote{https://github.com/argman/EAST} uses a stronger backbone (ResNet), RRD still performs better, achieving 2.6 percents F-measure improvement. Moreover, using multi-scale inputs including $384 \times 384, 384 \times 768, 768 \times 384, 768 \times 768, 1024 \times 1024, 1536 \times 1536$, ``RRD+MS'' further achieves 11.3 percents improvement than RRD using single-scale input. Some qualitative results of RRD on this dataset are given in Fig.~\ref{fig:text_results_visu}.
\vspace{-4mm}
\paragraph{MSRA-TD500.}
MSRA-TD500 is also a line-based dataset which contains Chinese and English. Thus, the inception block is applied for this dataset. As shown in Tab.~\ref{tab:text_benchmarks}, RRD outperforms the state-of-the-art methods in terms of precision, recall, and F-measure. To further verify the effectiveness of rotation-sensitive regression, we also evaluate the ``Baseline+inc'' architecture, which degrades the performance of RRD by 5 percents in terms of F-measure.
\vspace{-4mm}
\paragraph{COCO-Text.}
The experiments on COCO-Text Challenge prove the accurate regression  of RRD. As shown in Tab.~\ref{tab:coco-text}, COCO-Text Challenge provides a more strict evaluation protocol in which the IOU threshold is set to $0.75$, where RRD has a huge superiority because the detection results of RRD is much more accurate, benefiting from the rotation-sensitive regression. Specifically, RRD achieves comparable results with the previous state-of-the-art method when the IOU threshold is set to $0.5$ while outperforms all the previous methods by at least 6.1 percents when the IOU threshold is set to $0.75$.

\begin{table}[ht]
\footnotesize
\centering
\begin{tabular}{|c|c|c|c|c|c|c|}
\hline
\multirow{2}{*}{Methods} & \multicolumn{3}{c|}{IOU=0.5}                     & \multicolumn{3}{c|}{IOU=0.75}                    \\ \cline{2-7} 
                         & R         & P      & F      & R         & P      & F      \\ \hline
UM~\cite{coco-text-challenge}                       & \textbf{0.66} & 0.48          & 0.55          & 0.31          & 0.23          & 0.26          \\ \hline
TDN SJTU v2~\cite{coco-text-challenge}               & 0.54          & 0.62          & 0.58          & 0.28          & 0.32          & 0.30          \\ \hline
Text Detection DL~\cite{coco-text-challenge}         & 0.62          & 0.60          & \textbf{0.61} & 0.26          & 0.25          & 0.25          \\ \hline
RRD+MS                   & 0.57          & \textbf{0.64} &  \textbf{0.61}           & \textbf{0.34} & \textbf{0.38} & \textbf{0.36} \\ \hline
\end{tabular}
\caption{Experimental results on COCO-Text Challenge.}
\label{tab:coco-text}
\vspace{-4mm}
\end{table}
\vspace{-4mm}
\paragraph{IC15.}
For the comparison of RRD with the state-of-the-art results on IC15 dataset, we use the scale of $1024 \times 1024$ for single scale testing in RRD. The multi-scale testing includes the scales of $384 \times 384$, $768 \times 768$, $1024 \times 1024$, and $1536 \times 1536$. The comparison with the state-of-the-art results on IC15 dataset is given in Tab.~\ref{tab:text_benchmarks}. Even though IC15 is a word-based dataset whose text bounding boxes are not extremely long, RRD still achieves about 1 percent performance gain than the baseline with both single scale and multi-scale settings. Furthermore, RRD outperforms the state-of-the-art results by 4 percents with single scale setting and 2.8 percents in the case of multi-scale setting.

\vspace{-4mm}
\paragraph{IC13.}
\begin{table}[ht]
\footnotesize
\centering
\begin{tabular}{|c|c|c|c|}
\hline
Method                       & Recall & Precision & F-measure \\ \hline
FCRNall+filts~\cite{SynthText}  & 0.76            & 0.92        & 0.83            \\ \hline
TextBoxes~\cite{LiaoSBWL17}                    & 0.74   & 0.88      & 0.81      \\ \hline
TextBoxes+MS ~\cite{LiaoSBWL17}                & 0.83   & 0.89      & 0.86      \\ \hline
Seglink~\cite{seglink} & 0.83            & 0.88        & 0.85            \\ \hline
Tian et al.~\cite{eccv/TianHHH016}  & 0.83            & \textbf{0.93} & 0.88            \\ \hline
Tang et al.~\cite{tip/TangW17}     & 0.87      & 0.92 & \textbf{0.90} \\ \hline
He et al.~\cite{He_2017_ICCV} & 0.86      & 0.89 & 0.88 \\ \hline
He et al.~\cite{deepdirect} & 0.81      & 0.92 & 0.86 \\ \hline
WordSup+MS~\cite{Hu_2017_ICCV} & \textbf{0.88}      & 0.93 &\textbf{0.90}  \\ \hline
Baseline    & 0.74   & 0.88      & 0.81      \\ \hline
Baseline+MS   & 0.85   & 0.92      & 0.88      \\ \hline
RRD    & 0.75   & 0.88      & 0.81      \\ \hline
RRD+MS & 0.86   & 0.92      & 0.89      \\ \hline
\end{tabular}
\caption{Experimental results on IC13. MS stands for multi-scale testing.}
\label{table:IC13}
\end{table}

Although the proposed RRD is specifically designed for multi-oriented text detection, we also evaluate RRD on IC13 dataset consisting of horizontal text. The experimental results are depicted in Tab.~\ref{table:IC13}, showing that the proposed scheme has no performance loss in horizontal text detection. RRD achieves comparable performance with the state-of-the-art results on IC13. Moreover, compared with the most related work~\cite{LiaoSBWL17}, RRD also performs better.

\subsection{Results on the Ship Collection Benchmark}
There are three level tasks in HRSC2016 dataset.
Level 1 is to detect ship from backgrounds. Level 2 is to further give ship categories (\textit{war craft, aircraft carrier etc.}) and Level 3 steps forward to get ship types (\textit{car carrier, submarine etc.}). Basically, Level 1 task is enough to show our methods' superiority on multi-directional objects detection, so we compare our models with other carefully designed models on Level 1 task.

\begin{figure}
\centering
\includegraphics[width=0.95\linewidth]{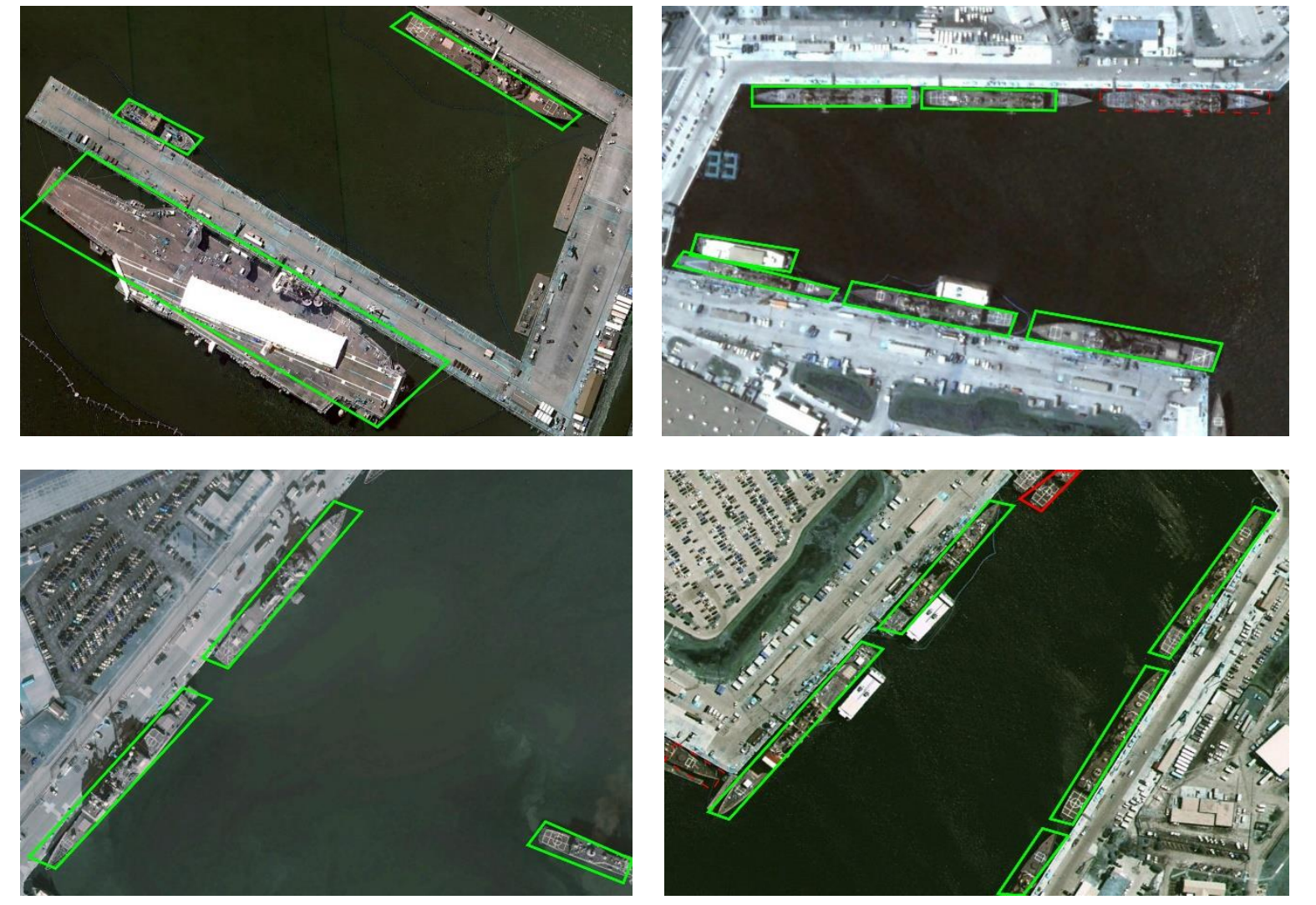}
\caption{Visualization of some results on HRSC2016. }
\label{fig:ship_vis}
\vspace{-4mm}
\end{figure}

Some visualization results are displayed in Fig.~\ref{fig:ship_vis}. 
As we can see, RRD accurately detects ships with arbitrary orientations.
Quantitative results are shown in Tab.~\ref{tab:ship datasets}. Instead of fine-tuning the model based on a pre-trained model, RRD is trained from scratch. Even so, RRD easily surpasses others with a promotion around 8.6 points in mAP reaching peak at 84.3~\cite{zk2017rotated}. Thus, it is obvious that RRD is not only suitable for scene text detection, but also skilled in other oriented object detection.

\begin{table}[ht]
\footnotesize
\centering
\begin{tabular}{|c|l|}
\hline
Methods                   & mAP           \\ \hline
Fast-RCNN~\cite{fast_rcnn}+SRBBS          & 55.7          \\ \hline
Fast-RCNN~\cite{fast_rcnn}+SRBBS+RBB      & 69.6          \\ \hline
Fast-RCNN~\cite{fast_rcnn}+SRBBS+RBB+RRoI & 75.7          \\ \hline
RRD(Ours)                     & \textbf{84.3} \\ \hline
\end{tabular}
\caption{Experimental results on HSRC2016. SRBBS(Ship Rotated Bounding Boxes Space) means labeling of ships is rotated polygon; RBB(Rotated Bounding Boxes) extends Fast-RCNN to a method capable of regressing rotated bounding boxes; RRoI(Rotated Region of Interest) pooling layer is implemented based on original RoI pooling, which is specially designed for rotated bounding boxes. For detailed description, refer to ~\cite{zk2017rotated}}
\label{tab:ship datasets}
\vspace{-4mm}
\end{table}

\subsection{Limitations}

\begin{figure}[htbp]
\begin{center}
\subfloat[\label{subfig-1:false_results}]{%
       \includegraphics[width=0.2\textwidth]{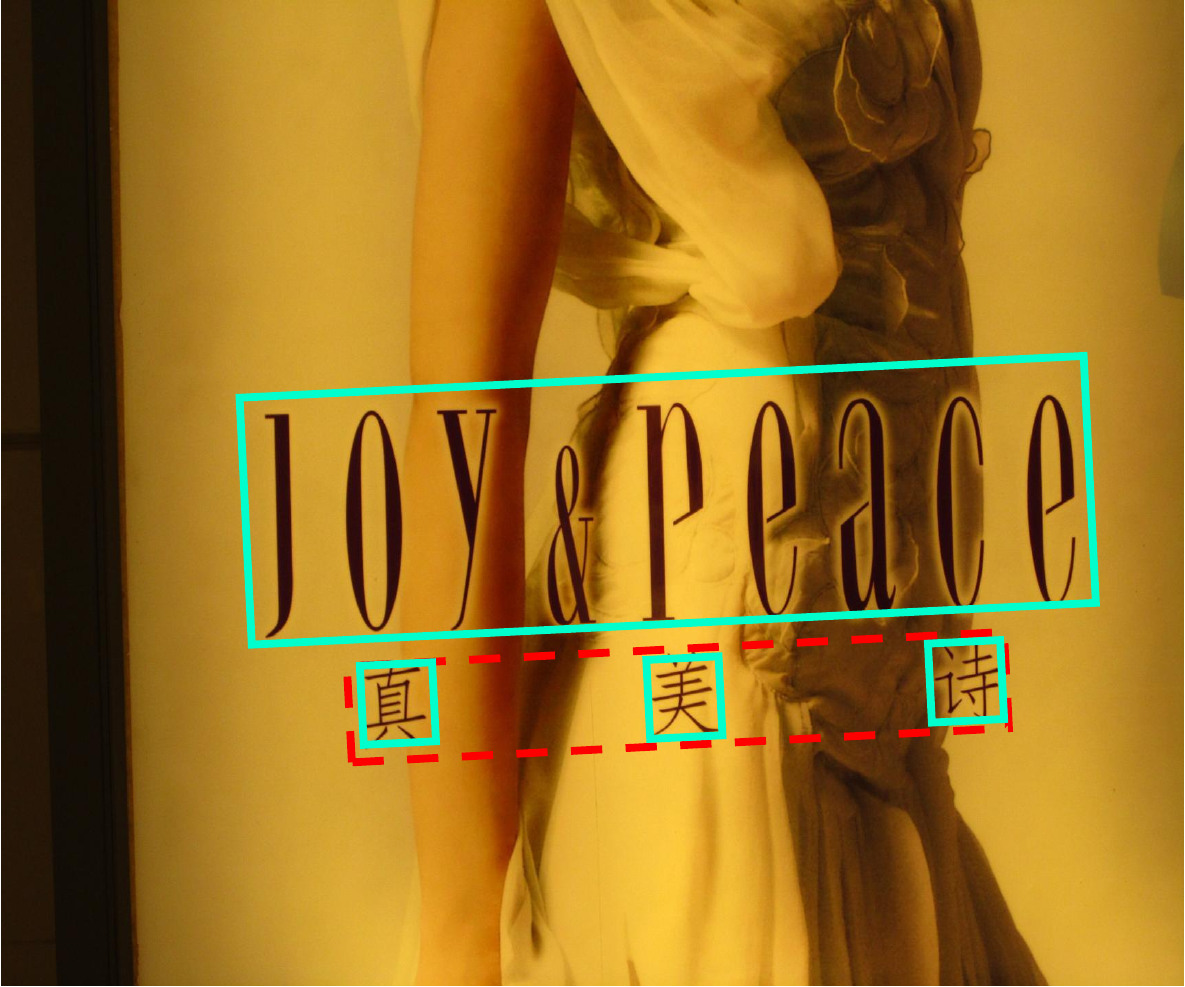}
     }
\subfloat[\label{subfig-2:false_results}]{%
       \includegraphics[width=0.2\textwidth]{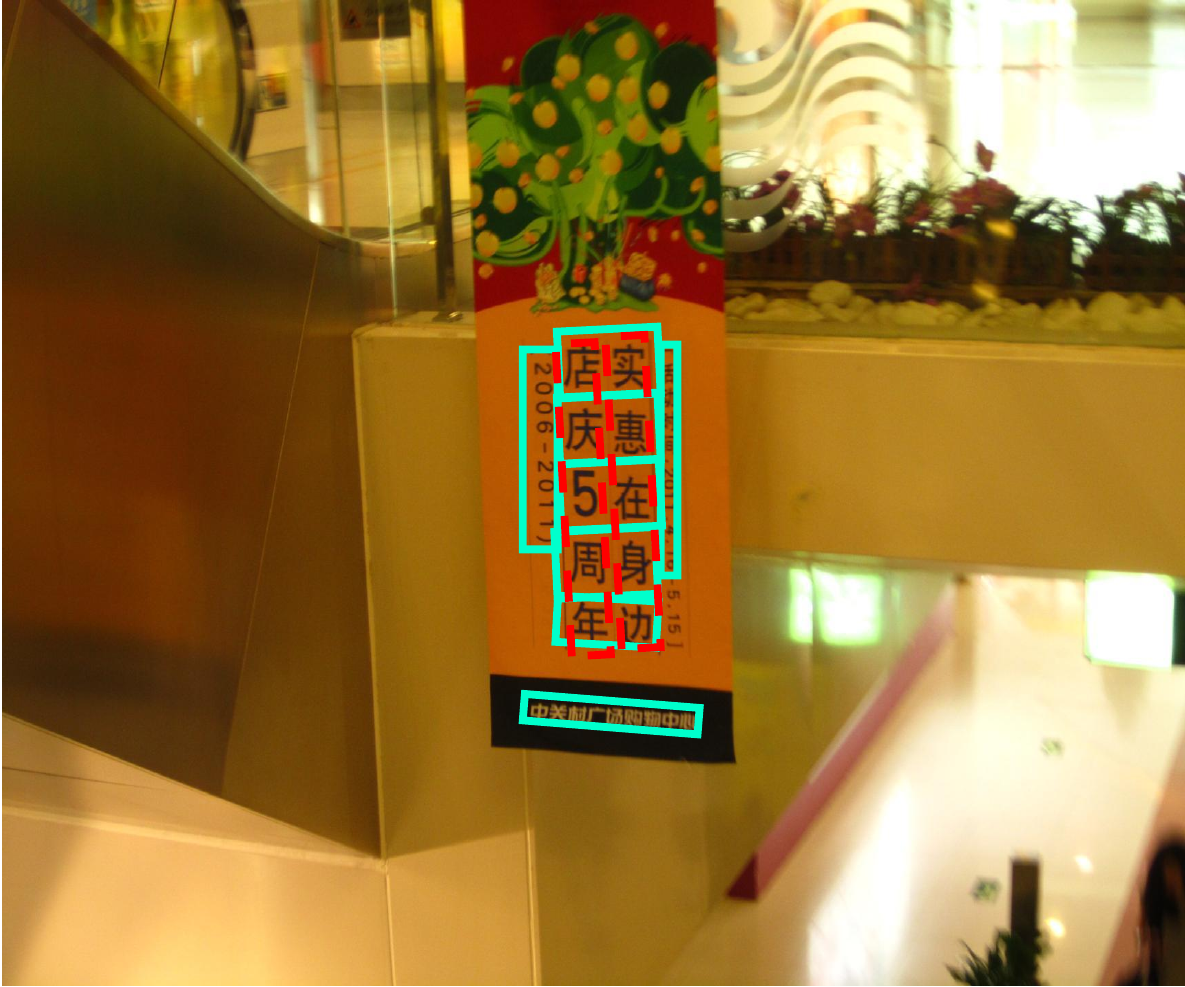}
     }
\end{center}
\caption{Visualization of limited occasions. Green bounding boxes: outputs of RRD; Red dashed bounding boxes: missing ground truths.}
\label{fig:false_results}
\vspace{-4mm}
\end{figure}

We observe that RRD fails to detect certain types of text.
As shown in Fig.~\ref{subfig-1:false_results}, RRD fails to detect a whole bounding box for a text line with large character spacing.
Another failure case is shown in Fig.~\ref{subfig-2:false_results}, where RRD incorrectly detects two vertical text lines as multiple horizontal ones.
We believe that detecting such text is beyond the current capability of RRD, also many other state-of-the-art methods.
Higher-level semantic understanding and spatial analysis may be required to address this issue.

\section{Conclusion} \label{sec:conclusion}
We have proposed RRD, a novel text detector that perform classification and regression using rotation-insensitive and sensitive features respectively.
This strategy is conceptually simple, yet its effectiveness and generality is well demonstrated on multiple datasets and tasks.
RRD improves the dense prediction of text presence and offsets, which can be found in many modern text detectors.
Potentially, these detectors will also benefit from the strategy we have adopted in RRD.
In the future, we are interested in stronger rotation-sensitive features and rotation-invariant features to further improve oriented object detection.
Also, since the principle of RRD goes beyond text detection and ship detection, we are interested in further exploiting its potentials in detecting other oriented objects such as those frequently appeared in aerial images~\cite{dota}.  

\section{Acknowledgements}
This work was supported by NSFC 61733007 and 61573160, to Dr. Xiang Bai by the National Program for Support of Top-notch Young Professionals and the Program for HUST Academic Frontier Youth Team.

{\small
\bibliographystyle{ieee}
\bibliography{egbib}
}

\end{document}